\documentclass[remotesensing,article,submit,moreauthors,pdftex]{Definitions/prev} 
\firstpage{1} 
\pubvolume{1}
\issuenum{1}
\articlenumber{0}
\pubyear{2021}
\copyrightyear{2021}
\datereceived{} 
\dateaccepted{} 
\datepublished{} 
\hreflink{https://doi.org/} 

\usepackage{pifont,amsfonts,amssymb,bbm,graphicx,bm} 

\renewcommand{\thefootnote}{\ding{\numexpr191+\value{footnote}}}

\Title{A Real-time Low-cost Artificial Intelligence System for Autonomous Spraying in Palm Plantations}
\TitleCitation{A Real-time Low-cost Artificial Intelligence System for Autonomous Spraying in Palm Plantations}

\Author{Zhenwang Qin $^{1}$\orcidA{}, Wensheng Wang $^{1,}$*, Karl-Heinz Dammer $^{2}$, Leifeng Guo $^{1,}$*\orcidB{} and Zhen Cao $^{1}$}

\AuthorNames{Zhenwang Qin, Wensheng Wang, Karl-Heinz Dammer, Leifeng Guo and Zhen Cao}
\AuthorCitation{Qin, Z.W.; Wang, W.; Dammer, K.H.; Guo, L.; Cao, Z..}
\AuthorCitation{Qin, Z.W.; Wang, W.; Dammer, K.H.; Guo, L.; Cao, Z..}
\address{%
$^{1}$ \quad Agricultural Information Institute, Chinese Academy of Agricultural Sciences, No.12 Zhongguancun South St., Haidian District Beijing P.R.China; qinzhw@hotmail.com;wangwensheng@caas.cn;guoleifeng@caas.cn;caozhen02@caas.cn\\
$^{2}$ \quad Leibniz Institute for Agricultural Engineering and Bioeconomy, Department Engineering for Crop Production, Max-Eyth-Allee 100, 14469 Potsdam, Germany; KDammer@atb-potsdam.de}

\corres{Correspondence: wangwensheng@caas.cn, guoleifeng@caas.cn}

\preto{\abstractkeywords}{\nolinenumbers}

\abstract{In precision crop protection, (target-orientated) object detection in image processing can help navigate Unmanned Aerial Vehicles (UAV, crop protection drones) to the right place to apply the pesticide. Unnecessary application of non-target areas could be avoided. Deep learning algorithms dominantly use in modern computer vision tasks which require high computing time, memory footprint, and power consumption. Based on the Edge Artificial Intelligence, we investigate the main three paths that lead to dealing with this problem, including hardware accelerators, efficient algorithms, and model compression. Finally, we integrate them and propose a solution based on a light deep neural network (DNN), called Ag-YOLO, which can make the crop protection UAV have the ability to target detection and autonomous operation. This solution is restricted in size, cost, flexible, fast, and energy-effective. The hardware is only 18 grams in weight and 1.5 watts in energy consumption, and the developed DNN model needs only 838 kilobytes of disc space. We tested the developed hardware and software in comparison to the tiny version of the state-of-art YOLOv3 framework, known as YOLOv3-Tiny to detect individual palm in a plantation. An average F1 score of 0.9205 at the speed of 36.5 frames per second (in comparison to similar accuracy at 18 frames per second and 8.66 megabytes of the YOLOv3-Tiny algorithm) was reached. This developed detection system is easily plugged into any machines already purchased as long as the machines have USB ports and run Linux Operating System.}

\keyword{Object detection, Precise spraying, Embedded AI, Palm plantation, YOLO, NCS2}

\begin{document}

\section{Introduction}

Target-oriented crop protection means spraying pesticides on specific areas or objects. Especially in tree crops in orchards, there is a certain distance from one tree to another. Especially in plantations in tropical areas, the space between the objects to be treated can vary extremely. At present, it is common to spray the whole area uniformly with pesticides by using air-crafts especially in the rugged environment or ground equipment. 

In the past few years, battery-powered Unmanned Aerial Vehicles (UAV) have been playing more and more important roles in agriculture. UAVs allow farmers to increase efficiency in certain aspects of the farming process. The crop protection drones carrying spray equipment are developed\cite{yang2018application, xiongkui2017recent, lan2018current} to spray crops in small fields or rugged environments such as mountain areas. In precise target-orientated pesticide spraying, drones can perform selective spraying of certain areas leaving our non-target areas. The two leading manufacturers, DJI\footnote{\url{https://www.dji.com}} and XAG \footnote{\url{https://www.xa.com}}, have released dozens of thousands of these drones to market in China. They have achieved big success, especially in field crops. 

However, applications in the orchard are still hindered by the complicated environments such as the irregularity of the canopy or the variety of height of trees. A similar situation occurs in palm plantations in tropical areas. Palms are often varying in heights. Older plants are usually more than dozens of feet in height. This complicated condition often causes drone crashes, so that requiring high manipulation skills. Furthermore, especially in tropical areas, the space between the palms can be largely varied. If pesticides would be applied uniformly like in present cases, the economic and ecological losses of pesticides are enormous. The spray liquid can hit the ground between the palms. Endowing a drone with computer vision (CV) function by an on-board camera and an embedded system is a valid way to solve this problem. This paper described equipment that can recognize the palm cones automatically in real-time while flying over the plantation.

In palm plantations, a practical computer vision module must be endowed to the UAVs. UAVs are designed in such a way to be able to detect plants via an on-device camera and react in real-time autonomously during a flight instead of controlled by the UAV pilot. For CV tasks, DNNs have shown superior performance and significantly outperformed other existing approaches due to their robustness to the diversity of targets. However, the powerful representation of DNNs often comes with a high computation complexity and memory demand, most of which are from the convolutional layers. This is not a problem in high-end Graphics Processing Unit (GPUs), but for a resource-constrained device such as a drone, we need to balance the DNN accuracy and hardware performance with tight constraints on the computational power, memory size, energy consumption, and latency.

The object detection algorithm "R-CNN"\cite{girshick2014rich} firstly introduced the DNN method in detecting objects from images. Combined with Selective Search\cite{uijlings2013selective}, R-CNN achieved notable improvements in comparison to traditional methods. Since then, there had been lots of research interests in applying CNN in CV tasks such as object detection and segmentation. 

The R-CNN series(\cite{girshick2014rich,girshick2015fast,dai2016r,ren2016faster,he2017mask}) algorithms intuitively divide an object detection task into two stages: 1) using DNN to extract features from an image and at the same time generate large numbers of region proposals that would contain objects of interest, 2) detection components classify these proposals to retrieve their categories and perform position regression to located objects precisely. Hence are called "two-stage" detectors. 

Although the accuracy has been demonstrated in many research, region proposal modules require huge computation and run-time memory footprint, thus making detection relatively slow, even in high-end GPUs. 

To improve the detection speed, some approaches are called one-stage detectors, represented by the "You Only Look Once (YOLO)" series(\cite{redmon2016you, redmon2017yolo9000, farhadi2018yolov3, bochkovskiy2020yolov4}), "Single Shot MultiBox Detector (SSD)" \cite{liu2016ssd}, CornerNet\cite{law2018cornernet} and CenterNet\cite{duan2019centernet} frame object detection as a regression problem to spatially separated bounding boxes and associated class probabilities. Encapsulating all the computations in a single network, single-stage detectors are more likely to run faster than two detectors while maintaining similar accuracy. Moreover, as the whole detection pipeline is a single network, a one-stage detector can be optimized end-to-end directly.

Among these single-stage detectors, YOLO series models are the fastest with state-of-art detection accuracy and hence become one of the most popular deep object detectors in practical applications. In this paper, we adopt YOLOv3-tiny, a compact version of YOLOv3 as the baseline. 

A YOLO series detector consists of three components, 1) a backbone DNN which extracts features from images, 2) one or more necks which are layers inserted between the backbone and head to collect feature maps from different stages, 3) multiple heads which are used to predict classes and bounding boxes.

Although some authors argue that the YOLOv3 series cannot provide the best accuracy in terms of small objects\cite{xu2020automated}, however, in a UAV-based platform, especially for guiding a drone to perform tasks such as spraying, the objects in the scene are usually not that small. On the other hand, we argue that the resolution of a detector can be improved by using a better backbone network, and prove that in our experiment.

In this research, we deployed a real-time object-detector to an embedded system attached to a UAV platform. On-device computing is very challenging since embedded devices usually have tight constraints on computational power, memory size, and energy consumption, which prohibits the use of the complex state-of-art network. Recent researches show three possible approaches to deal with this challenge, 
\begin{enumerate}

\item use an effective backbone DNN, such as SqueezeNet\cite{iandola2016squeezenet}, MobileNet v1\cite{howard2017mobilenets}, MobileNet v2\cite{sandler2018mobilenetv2}, ShuffleNet v1\cite{zhang2018shufflenet}, and ShuffleNet v2\cite{ma2018shufflenet};

\item use a hardware accelerator\cite{lee2018techology, li2020survey, mazzia2020real};

\item compress the dense model into sparse or low-bit architecture, such as weights quantization(\cite{rastegari2016xnor, zhou2016dorefa, wei2018quantization}), network pruning (\cite{wen2016learning, he2017channel, liu2017learning, gamanayake2020cluster}.

\end{enumerate}

\begin{figure}[H]	
\centering\includegraphics[width=10cm]{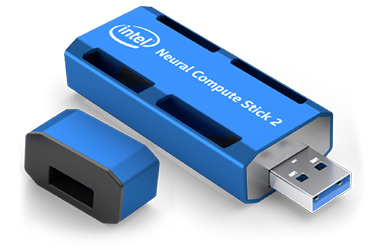}
\caption{The Intel NCS2 hardware.\label{fig1}}
\end{figure}

There are some embedded AI computing options including graphics processing units (GPUs), vision processing units (VPUs), and field-programmable gate arrays (FPGAs), and today commercial products are on the market such as nVIDIA, Intel, and MYiR. As listed in Table \ref{tab:tab1}, Intel NCS2 has both the least weight and least power consumption. In a battery-powered device, those features are important advantages, so this paper started from NCS2( Figure \ref{fig1}).
 
\begin{specialtable}[H] 
\caption{Main specifications of the candidate platforms in this work. Weight of NCS2 is not including the outer shell.\label{tab:tab1}} 

\begin{tabular}{p{3.1cm}p{2.9cm}p{4.8cm}p{6cm}}
\toprule & \textbf{Intel NCS2}	& \textbf{nVidia Jetson Nano}	& \textbf{MYiR ZU3EG}\\
\midrule
\textbf{Features Size}		& 73mm × 36mm			& 70mm × 45mm			& 100mm × 70mm\\
\textbf{HW Accelerator}		& Myriad X VPU			& 128-core nVidia Maxcell GPU			& Xilinx UltraScale\\
CPU		& N./A.			& Arm A57			& MPSoC XCZU3EG (4-core Arm A53)\\
\textbf{Peak Performance}		& 150GFLOPs			& 472GFLOPs			& 1.2TFLOPs\\
\textbf{Data Precision}		& FP16			& FP16/FP32			& FP32\\
\textbf{Nominal Power}		&1.5 Watts			& 10 Watts			& 10 Watts\\
\textbf{Weight}		&18 grams			& 140 grams			& 150 grams\\
\bottomrule
\end{tabular}
\end{specialtable}
 
In the following sections, direct matrices such as inference time or frames per seconds (fps) are used to evaluate a model, rather than computation volume theoretical indicator (BFLOPs). This paper seeks to get a model with the best accuracy under the constraint of real-time, which means a better speed performance of 24fps. 
This paper also demonstrates that a high-end GPU workstation is not always a prerequisite in training a real-life DNN model, even from scratch. During the research, a laptop equipped with an Intel i7-8750H CPU and an nVidia GTX1060 GPU, meanwhile equipped with 16GB system memory and 6GB GPU memory has been used. At the same time, the research evaluated different efficient DNN networks, and hereby further optimizing one of them was done to get the best real-time performance in a low-cost embedded device.

The developed system was tested in an \textit{Areca catechu L.}-plantation also known as betel-nut. This kind of very important high-value crop in tropical areas like India and other Southeast Asian countries. In the Hainan Island of China, this crop provides a livelihood to more than 2 million people in the rural area. This cultivar suffers from the Yellow Leaf Disease (YLD) caused by phytoplasma. It can lead to decay and then wilt of the palms\cite{menon1963yl}. This disease was firstly reported in India in 1949 and reached Hainan Island in 1984. To spray pesticides from the top of the palms is one effective way to control this disease.

\setcounter{section}{1}

\section{Related Works}
\subsection{Efficient DNN Architectures Optimized for Embedding Hardware}

Before the classic DNN Architecture named VGG16\cite{simonyan2014very} was proposed, backbones of DNN-based object detection frameworks often reuse image classifiers that are pre-trained by large datasets such as PASCAL, COCO, etc. They often have complex structures although they do have a very good performance on these datasets, however, there is no methodology to guide researchers to design their networks. VGG16 proposed the idea of building a deep model by reusing simple small 1×1 and 3×3 basic blocks. By stacking them and down-sampling blocks such as pooling twice or stride more than twice convolutional layers alternatively, VGG16 sets an example for a typical backbone network of an object detector, which can be divided into 5 stages, and finally outputs feature maps of 1/32 dimension of input dimensions, which provides enough information to be transformed to a prediction head.

VGG16 also demonstrated better accuracy can be achieved through a deeper network, but training a deep network is usually very challenging due to the vanishing gradient problem until He et al.\cite{he2016deep}. proposed Residual Network(a.k.a. ResNet). 

ResNet drives the deep learning researchers to pursue higher performance with new algorithms without consideration of parameter size, computer requirement, and inference time. Nevertheless, Iandola, Forrest N. et al.\cite{iandola2016squeezenet} stepped into another direction by decreasing the number of learnable parameters and reducing the amount of calculation of the entire network through its innovative "Fire module". This attempt increases the inference speed to the greatest extent although at the cost of greatly reducing the accuracy. Their work known as "SqueezeNet" achieves AlexNet-level accuracy on the ImageNet dataset with 50x fewer parameters. Additionally, with model compression techniques, SqueezeNet can be compressed to less than 0.5MB (510× smaller than AlexNet). 

Later, Howard et al.\cite{howard2017mobilenets} came up with MobileNet v1, reducing parameters and FLOPs dramatically without compromising accuracy by introducing depth-wise separable convolution. Architectures based on this idea can be used to perform image classification and object detection tasks on devices with limited resources.

Zhang, X. et al.\cite{zhang2018shufflenet} proposed ShuffleNet v1, which is designed especially for mobile devices with very limited computing power (e.g.,1.0-1.5 BFLOPs). The new architecture utilizes two new operations, point-wise group convolution and channel shuffle, to greatly reduce computation cost while maintaining accuracy. 

Sandler, M. et al.\cite{sandler2018mobilenetv2} argued that it is important to remove non-linearities in the narrow layers to maintain representational power, with the following 2 assumptions, 1) If the manifold of interest remains non-zero volume after ReLU transformation, it corresponds to a linear transformation. 2) ReLU is capable of preserving complete information about the input manifold, but only if the input manifold lies in a low-dimensional subspace of the input space And hereby proposed MobileNet v2. These assumptions are also proved somehow in our later experiments. 

Wang, R. J. et al.\cite{wang2018pelee} argued that models such as MobileNet v1\cite{howard2017mobilenets}, ShuffleNet v1\cite{zhang2018shufflenet}, and MobileNet v2\cite{ma2018shufflenet} are heavily dependent on depth-wise separable convolution which lacks efficient implementation in most deep learning frameworks and propose PeleeNet. PeleeNet achieves higher accuracy and over 1.8 times faster speed than MobileNet and MobileNetV2 on NVIDIA TX2 with only 66\% of the model size of MobileNet. However, we find that PeleeNet is much slower than the original YOLOv3-tiny architecture, it takes about 3 times longer to generate results, and also has 3 times bigger parameter size. Luckily enough, the neck part is found much more effective than the original one in YOLOv3-tiny architecture, so we use it to assemble our architecture, See Section 4 for details.

Ma, Ningning et al.\cite{ma2018shufflenet} firstly summarized two principles that should be considered for effective network architecture design in ShuffleNet v2. 1). The direct metric (e.g., speed) should be used instead of the indirect ones (e.g., FLOPs). 2). Such metrics should be evaluated on the target platform. These two principles are adopted in this work.

The backbone used in the version of YOLOv3-tiny does not refer to any of the above works. Instead, it utilizes a very straightforward version guideline by VGG16\cite{simonyan2014very}. Unlike full-size YOLOv3\cite{farhadi2018yolov3}, YOLOv3-tiny, the compact version of YOLOv3 detects objects out of 2 different scales feature maps, each one of them uses three prior boxes to predict. 

\subsection{Model Compression}

From a communication perspective, a DNN is somewhat similar to a communication channel that transforms information encoded in the input image, so redundancy is usually present in the networks brings unnecessary computation. When deploying a deep model on resource-limited devices, model compression is a useful tool for researchers to eliminate the redundancy, among which model pruning methods, i.e. structured pruning are widely used.

The key to structured pruning is to identify structured unimportance. Wen, W. et al.\cite{wen2016learning} propose a Structured Sparsity Learning (SSL) method to regularize the structures (i.e., filters, channels, filter shapes, and layer depth) of DNNs. They apply group Lasso to regularize multiple DNN structures and achieve both speedups of convolutional layer computation and improvement of classification accuracy. He, Yihui et al.\cite{he2017channel} propose an iterative two-step algorithm to effectively prune each layer of a trained CNN model, by a LASSO regression-based channel selection and least square reconstruction during test time. The challenge of their methods is extra work is required to prune a model, moreover, it’s difficult to implement LASSO regression. Based on the fact that in modern DNN, a convolutional layer is followed by a batch normalization layer to solve the problem of Internal Covariate Shift (ICS)\cite{ioffe2015batch}, Liu, Zhuang et al.\cite{liu2017learning} propose a method named Network Sliming to learn the channel-level structural importance by imposing an L1-Norm regularization on the $\gamma$ parameter of in the Batch Normalization (BN) layer. Network Slimming directly applies to modern CNN architectures, introduces minimum overhead to the training process, and requires no special software/hardware accelerators for the resulting models. Therefore, we adopt Network Slimming in our work.

\subsection{Real-time Object Detection in Agriculture}

Object detection is an essential task in autonomous agricultural applications. In the past few years, advances in the computer vision community have translated to provision (computer vision in agriculture), achieving state-of-the-art results with the use of DNNs for object detection and semantic image segmentation. For object detection tasks, Faster R-CNN and YOLOv3 are two main algorithms adopted in recent researches, although Mask R-CNN and SSD are also used in some other researches\cite{xu2020automated}. By referring to "Real-time", those researchers usually talk about getting results in seconds or less on a computer. Except for Mazzia, V. et al.\cite{mazzia2020real} implement a real-time apple detection system on the embedded system, they build a YOLOv3-tiny architecture with predictions across three different scales instead of 2 scales in the original version. On a weighted 280g embedded device nVidia’s Jetson AGX Xavier, they get a speed of 30 fps, with 19 watts power consumption. Meanwhile, they get 5 fps on NCS2, with less than 6 watts. We empirically argue that the better accuracy acquired in YOLOv3\cite{farhadi2018yolov3} mainly comes from the deeper backbone, although prediction on 1 more feature map may introduce some performance benefit, more attention should be paid to explore backbone network enhancement.

Hanwen Kang and Chao Chen\cite{kang2019fruit} developed a multi-function DNN named DaSNet-V2 to perform the vision sensing of the working environment in apple orchards. DaSNet-V2 adopts a one-stage detector architecture to perform the detection and instance segmentation of fruits. Meanwhile, a semantic segmentation branch is grafted to the network to segment branches in orchards, DaSNet-V2 was tested on an nVidia’s Jetson TX2 and took an inference time of 287ms.

\setcounter{footnote}{0}
\renewcommand{\thefootnote}{\fnsymbol{footnote}}

\section{Materials and Methods}

\subsection{Field Sites and Image Preparation}

\begin{figure}[H]	
\centering\includegraphics[width=18cm]{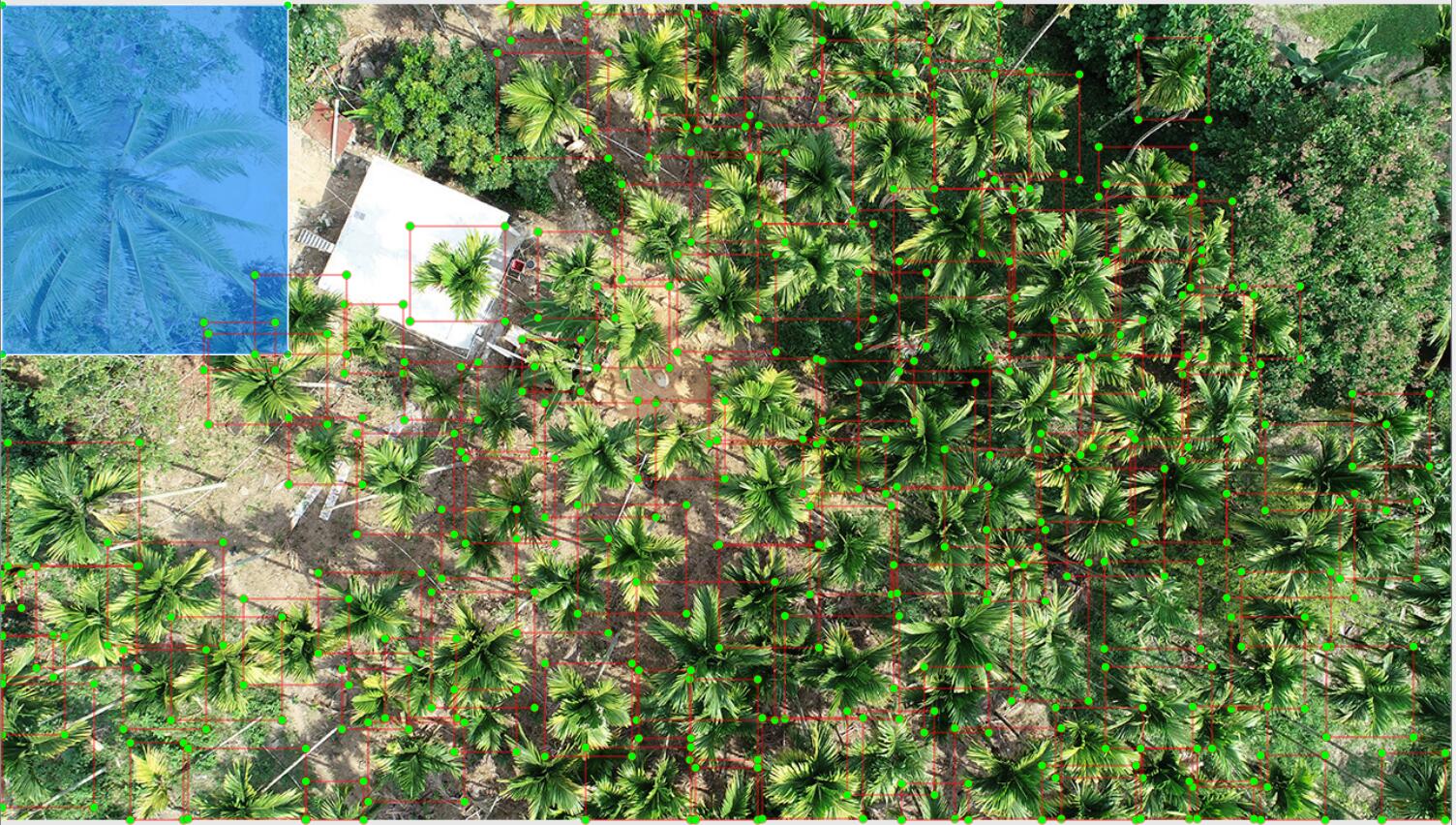}
\centering\caption{Areca palm plantation at the experimental site labeled with red quadrates. The blue semitransparent box covers a coconut palm (part of background)\label{fig2}}
\end{figure}

The \textit{Areca} plantations field sites are located near Sanya(18°15'10"N 109°30'42"E)\footnote{\url{https://en.wikipedia.org/wiki/Sanya}}, a city on the Hainan Island of China. The sites differed according to their age and their spatial distance from palm to palm. Some plantations are also heterogeneous regarding the individual trunk volume. Thus a high object variation was guaranteed. The images were collected by a DJI Phantom 4. An example image is shown in Figure \ref{fig2}. In total 400 images were taken at different times during the day and were labeled in an open-source software named labelImg, and randomly divided into two groups: one includes 300 images for training and the other includes 100 images for testing. 
The data augmentation method is the same as in YOLOv3\cite{farhadi2018yolov3}.

\subsection{Evaluation Metrics}

To evaluate the detection performance, inference time on NCS2, F1 score, and Intersection of Union (IoU) is applied. F1 score combines the performance evaluation of the recall and precision of the detection; hence it has been widely applied as the evaluation index in many previous studies of object detection when there is only one object category. The expression of the precision, recall, and F1 is presented as follow:

\begin{equation}
  P = \frac{TP}{TP+FP}
\end{equation}
\begin{equation}
  R = \frac{TP}{TP+FN}
\end{equation}
\begin{equation}
  F_1 = \frac{2\times P\times R}{P + R}
\end{equation}

\begin{figure}[H]	
\centering\includegraphics[width=10cm]{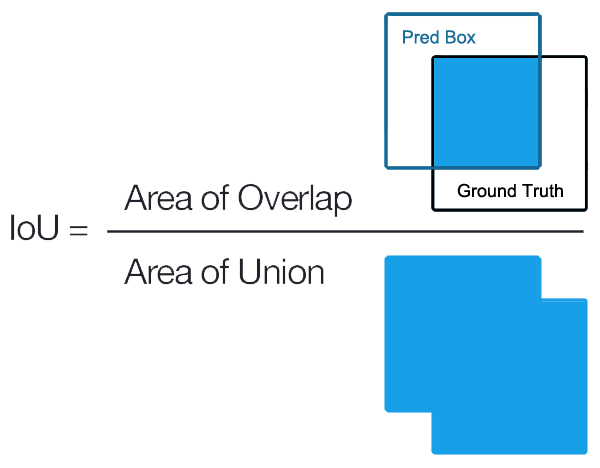}
\centering\caption{Definition of IoU.\label{fig3}}
\end{figure}

While $P$ denotes Precision, $R$ denotes recall, $TP$ denotes True Positives, $FP$ denotes False Positives, $FN$ denotes False Negatives.
The IoU is defined as in Figure \ref{fig3}. It measures the intersection area of the predicted object boundary box and the ground truth, to evaluate the location accuracy of the predicted boundary box of the prediction.

\subsection{Data Training}

Our frameworks are end-to-end trained on a single T1060 GPU, optimized by Adam\cite{kingma2014adam} and the initial learning rate is set to 0.001. Each mini-batch has 10 images so one epoch includes 15 mini-batches. We randomly resize the input dimension to (352, 352), (384, 384), (416, 416), (448, 448), (480, 480), (512, 512), (544, 544), (576, 576) for every epoch.
For every architecture, we train from scratch instead of using parameters from any pre-trained models. By adopting Leaky ReLU as an activation function and using Gaussian distribution initialized parameter, all the models are easily convergent in hundreds of thousands of iterations, which takes 2~4 days on an ASUS TUF Gaming FX86FM laptop. The value of Gaussian parameters $\mu = 0$, $\sigma = (16n)^{-0.5}$ while $n$ is the number of weight elements.

\textbf{L2-norm regularization} Regularization has been introduced to deep learning for a long time. It brings in additional information to prevent over-fitting. L2-norm regularization can be described as 
\begin{equation}
  L = \sum_{(x,y)}l(f(x,W),y)+\lambda\sum_{w}w^2
\end{equation}

In this equation, $\lambda$ is a super parameter, and in YOLO papers, it is referred to as "decay" or "weight decay" and set to 0.0005.
However, regularization takes into consideration in this work, and a model which eventually used on the NCS2 device had been trained as well. The whole flow is described as the following Figure \ref{fig4}:

\begin{figure}[H] 
\centering\includegraphics[width=12cm]{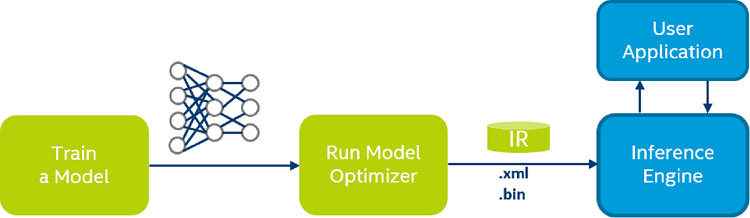}
\centering\caption{Network training and deployment on NCS2.\label{fig4}}
\end{figure} 

Some C++ tools for the training, testing, and model compressing were developed. During the training phase, the CUDA 11.1 and cuDNN 8.0 libraries are used, and all the parameters are in standard 32-bit float point values (FP32). However, NCS2 only supports 16-bit "half precision" float point values (FP16), which can express values in the range ±65,504, with the minimum value above 1 being 1 + 1/1024. To minimize accuracy loss during parameters being quantified from FP32 to FP16, parameters should be small enough. However, if we start with a small $\lambda$ value, we may get a model with a bunch of huge value parameters beyond FP16, especially in the first layer. Hence we set this value to 0.01 during the first 100k iterations and set it to 0.001 afterward.

\textbf{Loss} The YOLOv3\cite{farhadi2018yolov3} objector predicts bounding boxes using dimension clusters as prior boxes. For each bounding box there are 4 corresponding predicted values $t_x$, $t_y$, $t_w$, $t_h$, When the center of object is in the cell offset from the top left corner of the image by ($c_x$, $c_y$), and the prior box has dimension ($p_w$, $p_h$), then prediction values correspond to 
\begin{equation} 
\begin{split}
  b_x & = c_x + \sigma(t_x) \\ 
  b_y & = c_y + \sigma(t_y) \\ 
  b_w & = p_w + e^{t_w} \\ 
  b_h & = p_h + e^{t_h}
\end{split}
\end{equation}
As in Figure \ref{fig5}:
\begin{figure}[H] 
\centering\includegraphics[width=8cm]{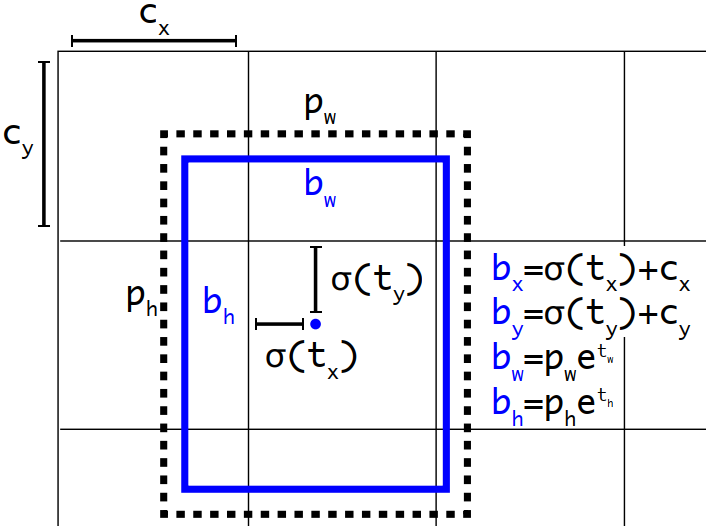}
\centering\caption{Bounding boxes with dimension priors and location prediction for YOLOv3.\label{fig5}}
\end{figure} 

For YOLOv3,prediction loss comprises 3 parts, the object loss $L_{obj}$, the classification loss: $L_{cls}$, and the coordinate loss $L_{box}$

\begin{equation}
  Loss = L_{obj} + L_{cls} + L_{box}
\end{equation}

Where

\begin{equation}
  L_{obj} = \lambda_{noobj}\sum_{i}^{S^2}\sum_{j}^{B}\mathbbm{1}_{i,j}^{noobj}(c_i-\hat{c_i})^2 + \lambda_{obj}\sum_{i}^{S^2}\sum_{j}^{B}\mathbbm{1}_{i,j}^{obj}(c_i-\hat{c_i})^2 
\end{equation}

\begin{equation}
  L_{cls} = \lambda_{cls}\sum_{i}^{S^2}\sum_{j}^{B}\mathbbm{1}_{i,j}^{obj}\sum_{c \in classes}p_i(c)log\left(\hat{p_i}(c)\right)\qquad\qquad\qquad\quad
\end{equation} 

\begin{equation}
\begin{split}
  L_{box} = \lambda_{box}\sum_{i}^{S^2}\sum_{j}^{B}\mathbbm{1}_{i,j}^{obj}\left(2-w_i \times h_i\right)\; \times \qquad\qquad\qquad\qquad\qquad\\ \left((x_i-\hat{x_i})^2+(y_i-\hat{y_i})^2+(w_i-\hat{w_i})^2+(h_i-\hat{h_i})^2\right)
  \end{split}
\end{equation}
 $S$ denotes the size of the feature map to be predicted, $B$ denotes the prior boxes count, $\mathbbm{1}_{i,j}^{obj}$ denotes that the $i$-th cell and the $j$-th prior box responsible for one ground truth, and $\mathbbm{1}_{i,j}^{noobj}$ denotes the opposite.

In this work, only palms need to be detected, hence always assume $L_{cls}=0$. Moreover, Focal Loss\cite{lin2017focal} is used in $L_{obj}$ to increase the recall rate $R$ and suppress the erroneous recall rate $FP$ :

\begin{equation}
  L_{obj} = -\lambda_{obj}\sum_{i}^{S^2}\sum_{j}^{B}\mathbbm{1}_{i,j}^{obj}\alpha\left(1-c_i\right)^{\gamma}\log(c_i)
\end{equation}
In this experiment, the parameters $\alpha$ and $\gamma$ had been set to 0.5 and 0.2 respectively. The intuition for Focal Loss is that the bigger the gap between prediction and ground truth, the more attention is paid to the prediction error, when one prediction value is big enough (bigger than the threshold, but still less than the target value(1.0), the loss value is quite small, sometimes we don't have to care for such loss too much.

In terms of $L_{box}$, CIOU Loss proposed in \cite{zheng2020distance} in used just as Bochkovskiy, Alexey et, al.\cite{bochkovskiy2020yolov4} do, as follows,
\begin{equation}
 L_{box} = \lambda_{box}\sum_{i}^{S^2}\sum_{j}^{B}\mathbbm{1}_{i,j}^{obj}\left(1-IoU_i+\frac{(x_i-\hat{x_i})^2+(y_i-\hat{y_i})^2}{c_i^2}+\frac{v_i^2}{(1-IoU_i)+v_i}\right)  
\end{equation}
Where $c_i^2$ is the area of the minimum box containing the prediction box and ground truth box. And 
\begin{equation}
  v_i = \frac{4}{\pi}\left(\arctan{\frac{\hat{w_i}}{\hat{h_i}}}-\arctan{\frac{w_i}{h_i}}\right)
\end{equation}
The values of $\lambda_{obj}$ and $\lambda_{box}$ are set to 1 and 0.2 respectively, but when the model is hard to convergence, $\lambda_{box}$ can be adjusted according to the condition.

\textbf{Network Slimming} Researches have demonstrated that accuracy can be improved through increasing layers (a deeper layers)\cite{simonyan2014very} or increasing channels in layers (a wider layers)\cite{howard2017mobilenets}. In this experiment, a wider and deeper enough initial network architecture had been used, and let the network learn its structural sparsity, and Network Slimming had been used as well which was introduced in the previous section. The slimming is performed on a well-trained network when the importance among the $\gamma$ parameters in the BN layers are further learned, no regularization is imposed on parameters in the convolutional layers, and after pruning, the model is re-trained.

\begin{figure}[H]	
\centering\includegraphics[width=10cm]{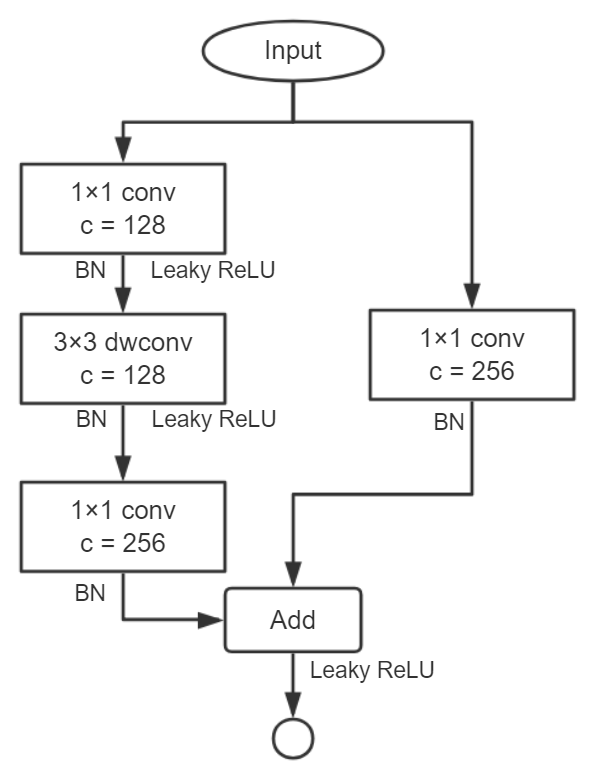}
\centering\caption{The ResBlock in PeeleNet.\label{fig6}}
\end{figure}

The training scheme in network slimming is similar to normal training, $\lambda$(the weight\_decay value) starts from 0.01 and then 0.001 after 100k iterations. 
An interesting observation is when replacing the head used in YOLOv3-Tiny with the ResBlock proposed in PeeleNet\cite{wang2018pelee}, as shown in Figure \ref{fig6}, and the left branch of ResBlock designed as Inverted Residuals Bottlenecks defined in MobileNet v2\cite{sandler2018mobilenetv2} and set expansion factor, weights in some channels in the first 1×1 convolutional layer and second 3×3 convolutional layer are all trends to zero, that is, the structure finally become very similar to a standard ResBlock. 

\textbf{Background Training} 
As shown in Figure \ref{fig2} in most areca plantations, the contrast between foreground and background was not obvious. Almost all the image is green with some variation from light to dark green with some yellowish-brown spots, which were the YLD diseased palm individuals. To consolidate the training, we label all the palms in the dataset and hence we can train the predictor to avoid wrong predicting background as the object to increase accuracy.
We define those prior boxes, which do not overlap with any of the ground truths as "background boxes", and punish prediction error of these boxes, make sure the confidence is less than the threshold (say, 0.5) to decrease false-positive predictions.

\subsection{Prior Boxes}

Some studies have dug into the selection of prior boxes in the YOLO model. It is empirically believed that some loss of accuracy comes from the unequal distribution of the ground truth by anchors, in another word, one specific prior box in a cell response to predict more than the ground truth so that during the training process, there is no way to learn all the ground truth. One way to solve this problem is to avoid this conflict, for example, use better designed prior boxes array or a bigger prior box collection. There is no necessary to use more predictors for a light model (i.e. in \cite{mazzia2020real}) if the backbone network has well enough representational power since more predictors bring more computation complexity. In our experiment, we stick to this argument and hereby run k-means over our datasets to pick prior boxes for our model, as in YOLOv3-tiny, we first set k=6 and get a box array of (23,23), (35,36), (48,49), (64,66), (90,91), (147,157), referred to as "def-anchors" in the later section, with all the images are normalized to (416,416). Since the smallest box are bigger than a high-resolution cell grid(16,16) in both width and height, we use another box array of (10,14), (27,23), (37,58), (75,64), (93,104), (187,163), referred to as "cust-anchors" to see what happens if there is one smaller prior box than the smaller cell grid. We also use k=8 to get another box array of (19,19), (27,29), (37,36), (43,48), (58,57), (71,75), (99,101), (158,169), referred to as "8-anchors".

\subsection{Model Deployment to NCS2}

A well-trained model is compressed to FP16 data format and transformed according to the guidance of OpenVINO and tested on NCS2. The source code is integrated into our software and another tool is designed for testing, which is also available at \url{https://github.com/rossqin/}.  

\section{Results}

\begin{figure}[H]	
\begin{center}
\includegraphics[width=12cm]{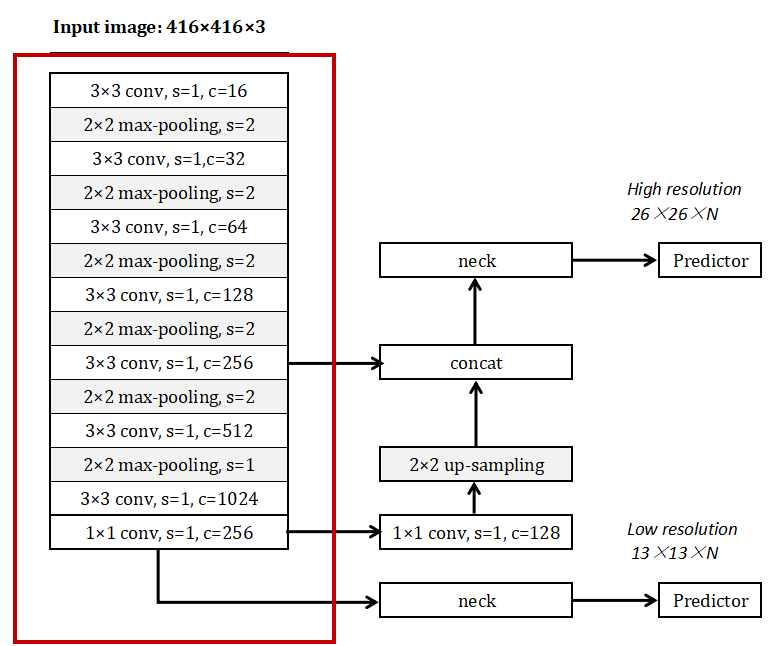}
\caption{The YOLOv3-tiny framework, s: stride, c: channel.\label{fig7}}
\end{center}
\end{figure}

The basic YOLOv3-tiny architecture is showed in Figure \ref{fig7}. and in the red frame, it’s darknet18. The firstly investigated results were brought into by different schemes, and then perform slimming on the best model and evaluate its performance on NCS2.

\subsection{Darknet18}

The darknet18 backbone is very simple and can be seen as the simplest structure following the principle proposed by VGG16, it is also a very good structure to inspect the performance data for NCS2. Afterward further replacing the neck with the one come up with PeleeNet. 

Table \ref{tab:tab2} shows increase-mental accuracy improvement by background training and prior boxes selection.

\begin{specialtable}[H] 
\caption{Accuracy increamental improvement.\label{tab:tab2}} 

\begin{tabular}{p{5cm}p{2cm}p{1.5cm}p{1.5cm}ccc}
\toprule
\multirow{2}{*}{\textbf {Model}} &
  \textbf {Average} & 
  \multicolumn{2}{c}{\textbf {F1 Score}} & 
  \textbf {Parameters} & 
	\multirow{2}{*}{\textbf {BFLOPs}} & 
	\textbf {Inference} \\
	\cline{3-4}
	&\textbf{IoU}& $\mathbf{IoU_{0.5}}$ & $\mathbf{IoU_{0.75}}$ & \textbf{Size} & & \textbf{Time on NCS2}\\
\midrule
\textbf{default-anchors-no-bg} &0.8186 &0.9236 & 0.7358 &8.634M &5.436 & 37.75ms \\
\textbf{default-anchors-bg} &0.8202 &0.9250 & 0.7503 &8.634M &5.436 & 37.75ms \\
\textbf{custom-anchors-bg} &0.8170 &0.9278 & 0.7422 &8.634M &5.436 & 37.75ms \\
\textbf{8-anchors-bg} &\textbf{0.8266} &\textbf{0.9300} & \textbf{0.7685} &8.668M &5.449 & 37.75ms \\
\bottomrule
\end{tabular}
\end{specialtable}

By training background and use more prior boxes, the accuracy is improved with a very small computational overhead, which does not reflect in inference time on NCS2. The reason that hand-selected prior boxes do not bring notable accuracy changes may be that there are not many small objects in the trained or tested images.

What needs to be pointed out is that the "Inference Time on NCS2" refers to the net computation time spent on the device, not including image decode and data transfer.

\subsection{The Res-Block Component}

Table \ref{tab:tab3} shows the performance improvement brought by the Res-Block neck. In this comparison, 8 prior boxes are used in predictors(4 by each).

\begin{specialtable}[H] 
\caption{Accuracy improvement by ResBlock.\label{tab:tab3}} 

\begin{tabular}{p{5cm}p{2cm}p{1.5cm}p{1.5cm}ccc}
\toprule
\multirow{2}{*}{\textbf {Model}} &
  \textbf {Average} & 
  \multicolumn{2}{c}{\textbf {F1 Score}} & 
  \textbf {Parameters} & 
	\multirow{2}{*}{\textbf {BFLOPs}} & 
	\textbf {Inference} \\
	\cline{3-4}
	&\textbf{IoU}& $\mathbf{IoU_{0.5}}$ & $\mathbf{IoU_{0.75}}$ & \textbf{Size} & & \textbf{Time on NCS2}\\
\midrule
\textbf{YOLOv3-tiny built-in} & \textbf{0.8266} &0.9300 & 0.7685 & 8.668M & 5.449 & 37.75ms \\
\textbf{ResBlock} & 0.8247 &\textbf{0.9432} & \textbf{0.7777} & \textbf{6.848M} & \textbf{4.109} & \textbf{27.42ms} \\
\bottomrule
\end{tabular}
\end{specialtable}

The F1 score in both $\mathbf{IoU_{0.5}}$ and $\mathbf{IoU_{0.5}}$ markedly improve at a cost of a slight drop in average IoU, which is acceptable. Furthermore, the parameters reduce from 8.668 million to 6.848 million, and BFLOPs reduce from 5.449 billion to 4.109 billion, leading to an inference time shrink of more than 10 mill-seconds, about 27\% of the original value.

\subsection{Network Slimming}

\setcounter{footnote}{0}

In the darknet source code\footnote{\url{https://pjreddie.com/darknet/}}, L2-norm regularization imposed on weights with $\lambda=0.0005$. This is too small and can lead to big parameters beyond FP16. Starting from $\lambda=0.01$, we find that in some channels, all the parameters trend to zeros, hence we can remove those channels to reduce computation. However, by imposing L1-norm regularization on the $\gamma$ parameters in batch normalization layers, we get a better result. We prune all the channels of which $\left|\gamma\right| < 0.5$. Table \ref{tab:tab4} shows the pruned results.

\begin{specialtable}[H] 
\caption{Performance data by pruning.\label{tab:tab4}} 

\begin{tabular}{p{5cm}p{2cm}p{1.5cm}p{1.5cm}ccc}
\toprule
\multirow{2}{*}{\textbf {Model}} &
  \textbf {Average} & 
  \multicolumn{2}{c}{\textbf {F1 Score}} & 
  \textbf {Parameters} & 
	\multirow{2}{*}{\textbf {BFLOPs}} & 
	\textbf {Inference} \\
	\cline{3-4}
	&\textbf{IoU}& $\mathbf{IoU_{0.5}}$ & $\mathbf{IoU_{0.75}}$ & \textbf{Size} & & \textbf{Time on NCS2}\\
\midrule
\textbf{ResBlock} & 0.8247 &0.9432 & 0.7777 & 6.848M & 4.109 & 27.42ms \\
\textbf{Pruned} & 0.8237 &0.9433 & 0.7625 & 6.548M & 3.949 & 27.33ms \\
\bottomrule
\end{tabular}
\end{specialtable}

Channel slimming does not bring notable performance enhancement in terms of inference time, on the contrary, it induces a little degradation to IoU and F1 scores in this experiment, however, in some architecture it generates a smaller model. 

\subsection{SqueezeNet}

Table \ref{tab:tab5} shows the performance result when darknet18 is replaced by SqueezeNet. SqueezeNet has better performance in terms of IoU and a smaller parameter size, however, double computation complexity and inference time. 

\begin{specialtable}[H] 
\caption{Performance data of SqueezeNet.\label{tab:tab5}} 

\begin{tabular}{p{5cm}p{2cm}p{1.5cm}p{1.5cm}ccc}
\toprule
\multirow{2}{*}{\textbf {Model}} &
  \textbf {Average} & 
  \multicolumn{2}{c}{\textbf {F1 Score}} & 
  \textbf {Parameters} & 
	\multirow{2}{*}{\textbf {BFLOPs}} & 
	\textbf {Inference} \\
	\cline{3-4}
	&\textbf{IoU}& $\mathbf{IoU_{0.5}}$ & $\mathbf{IoU_{0.75}}$ & \textbf{Size} & & \textbf{Time on NCS2}\\
\midrule
\textbf{default-anchors-no-bg} &0.8186 &0.9236 & 0.7358 &8.634M &5.436 & 37.75ms \\
\textbf{SqueezeNet} & 0.8346 & 0.9304 & 0.7860 & 1.186M & 5.176 & 56.67ms \\ 
\bottomrule
\end{tabular}
\end{specialtable}

\subsection{MobileNet v2}

\begin{figure}[H] 
\centering\includegraphics[width=6cm]{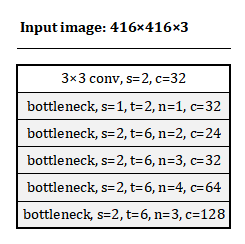}
\caption{The compact MobileNet v2 backbone.\label{fig8}}
\end{figure}

In the experiment, a compact version of MobileNet V2 had been used, the architecture is shown in Figure \ref{fig8}. bottleneck modules are as defined in the original paper. 

Table \ref{tab:tab6} shows its performance. With the bottleneck microstructure, improvements have been made in terms of IoU, F1 score, and BFLOP. However, the inference time is much slower (about 2.5 times of darknet18 takes).

\begin{specialtable}[H] 
\caption{Performance data of MobileNet v2.\label{tab:tab6}} 

\begin{tabular}{p{5cm}p{2cm}p{1.5cm}p{1.5cm}ccc}
\toprule
\multirow{2}{*}{\textbf {Model}} &
  \textbf {Average} & 
  \multicolumn{2}{c}{\textbf {F1 Score}} & 
  \textbf {Parameters} & 
	\multirow{2}{*}{\textbf {BFLOPs}} & 
	\textbf {Inference} \\
	\cline{3-4}
	&\textbf{IoU}& $\mathbf{IoU_{0.5}}$ & $\mathbf{IoU_{0.75}}$ & \textbf{Size} & & \textbf{Time on NCS2}\\
\midrule
\textbf{default-anchors-no-bg} &0.8186 &0.9236 & 0.7358 &8.634M &5.436 & 37.75ms \\
\textbf{MobileNet v2} & 0.8443 & 0.9592 & 0.8240 & 1.082M & 2.095 & 65.32ms \\ 
\bottomrule
\end{tabular}
\end{specialtable}

\subsection{ShuffleNet v2}

\begin{figure}[H] 
\centering\includegraphics[width=12cm]{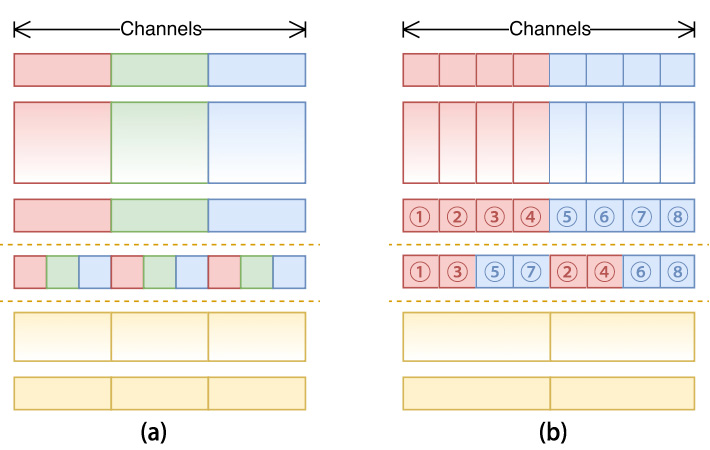}
\caption{(a) The Standard channel shuffle operation in ShuffleNet ; (b) channel reorganization operation in our work.\label{fig9}}
\end{figure}

ShuffleNet v2 building blocks are used to build our backbone except for some small modifications. There are two reasons. Firstly, the "Channel Shuffle" operation (see Figure \ref{fig9}.a) is not supported by the NCS2 hardware, so we use "Channel Reorganization" to achieve the same or similar effect, ( see Figure \ref{fig9}.b); Secondly, We find that In MobileNet v2, An activation layer follows a 3×3 depth-wise convolutional layer instead of a 1×1 convolutional layer and it brings a better accuracy. We compare the three occasions of 1) A 3×3 depth-wise convolutional layer followed by an activation layer and no activation layer for the 1×1 convolutional layer; 2) No activation layer for 3×3 depth-wise convolutional layers and a 1×1 convolutional layer is followed by an activation layer; 3) Both convolutional layers are followed by activation layers, details are described in Section 4.

\begin{figure}[H] 
\centering\includegraphics[width=11.6cm]{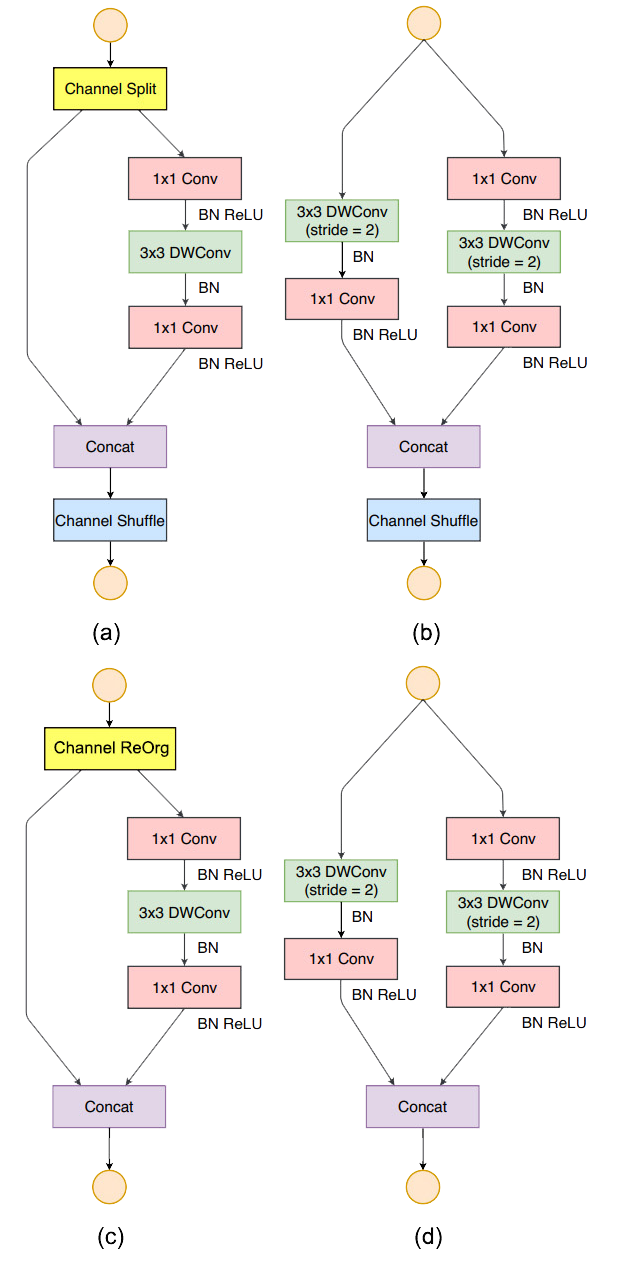}
\caption{Building Blocks of ShuffleNet v2 and this work. (a) the basic ShuffleNet v2 unit; (b) the ShuffleNet v2 unit for spatial down sampling (2×); (c) our basic unit; (d) our unit for spatial down sampling (2×).\label{fig10}}
\end{figure}

\begin{figure}[H] 
\centering\includegraphics[width=8cm]{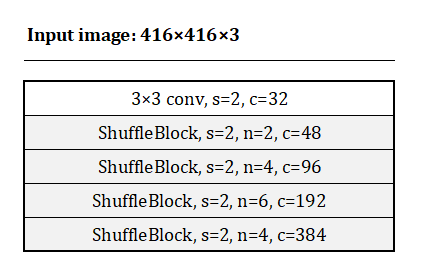}
\caption{The ShuffleNet v2 backbone. Each line describes a sequence of 1 or more identical (modulo stride) layers, repeated n times. All layers in the same sequence have the same number c of output channels. The first layer of each sequence has a stride s and all others use stride 1. All spatial convolutions use 3 × 3 kernels.\label{fig11}}
\end{figure}

Figure \ref{fig11} shows our backbone architecture. 

Table \ref{tab:tab7} shows the results. The second one has the best performance, and for the third one, more non-linearity leads to worse performance. 

Combine with a modified version of ShuffleNet-v2 backbone, a ResBlock neck, and a YOLOv3 head, we get Ag-YOLO. 

\begin{specialtable}[H] 
\caption{Performance data of ShuffleNet v2.\label{tab:tab7}} 

\begin{tabular}{p{5cm}p{2cm}p{1.5cm}p{1.5cm}ccc}
\toprule
\multirow{2}{*}{\textbf {Model}} &
  \textbf {Average} & 
  \multicolumn{2}{c}{\textbf {F1 Score}} & 
  \textbf {Parameters} & 
	\multirow{2}{*}{\textbf {BFLOPs}} & 
	\textbf {Inference} \\
	\cline{3-4}
	&\textbf{IoU}& $\mathbf{IoU_{0.5}}$ & $\mathbf{IoU_{0.75}}$ & \textbf{Size} & & \textbf{Time on NCS2}\\
\midrule
\textbf{default-anchors-no-bg} &0.8186 &0.9236 & 0.7358 &8.634M &5.436 & 37.75ms \\
\textbf{ShuffleNet v2(1)} & \textbf{0.8349} & 0.9448 & \textbf{0.7893} & 813K & 1.033 & 26.23ms \\ 
\textbf{ShuffleNet v2(2)} & 0.8278 & \textbf{0.9513} & 0.7668 & \textbf{711K} & \textbf{0.985} & \textbf{25.96ms} \\ 
\textbf{ShuffleNet v2(3)} & 0.8178 & 0.9404 & 0.7394 & 878K & 1.071 & 27.60ms \\ 
\bottomrule
\end{tabular}
\end{specialtable}

Ag-YOLO takes the second one, "No activation layer for 3×3 depth-wise convolutional layers and a 1×1 convolutional layer is followed by an activation layer".

\subsection{Models tested on NCS2}

All models are converted to OpenVINO-version and tested on the NCS2 device. The host is a Windows 10 laptop and data transferred via USB3 protocol which is also supported by the Raspberry Pi 4 computer. Due to the data precision loss, performance degradation is seen among all of the models. And from table \ref{tab:tab8}, we can see Ag-YOLO improves the original YOLOv3-tiny version significantly both in speed(about 6 frames more in a second) and accuracy( about 0.2 increase in F1 score). The model using a compact MobileNet v2 backbone surpasses our model a little in terms of F1 score and IoU, however takes double times longer to run.

When the input dimension is $352 \times 352 $, Ag-YOLO achieves the speed of 36.5fps, with an F1 score of 0.9205 and IoU of 0.708 on NCS2, while YOLOv3-tiny achieves similar accuracy at the speed of 18.1 fps, when the input dimension increases to $448 \times 448 $. 

Based on these data, Ag-YOLO is two times faster than YOLOv3-tiny.

\begin{figure}[H] 
\centering
\includegraphics[width=9cm]{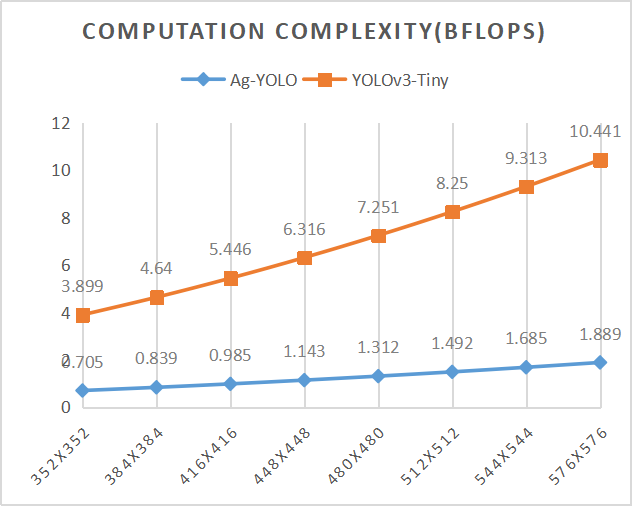}\quad \includegraphics[width=9cm]{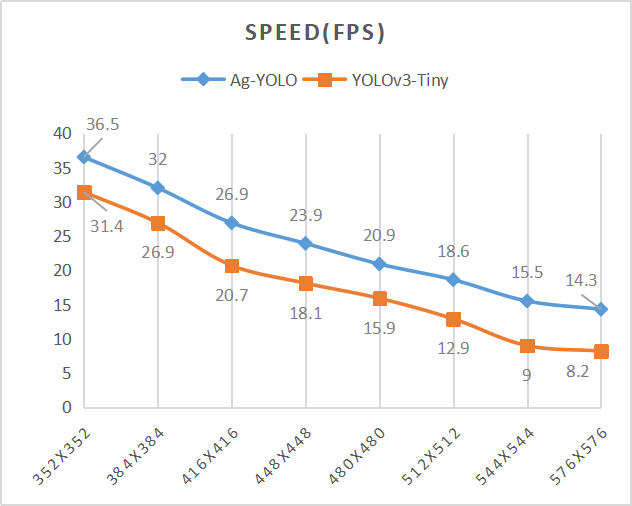}
\\[6pt]
\includegraphics[width=9cm]{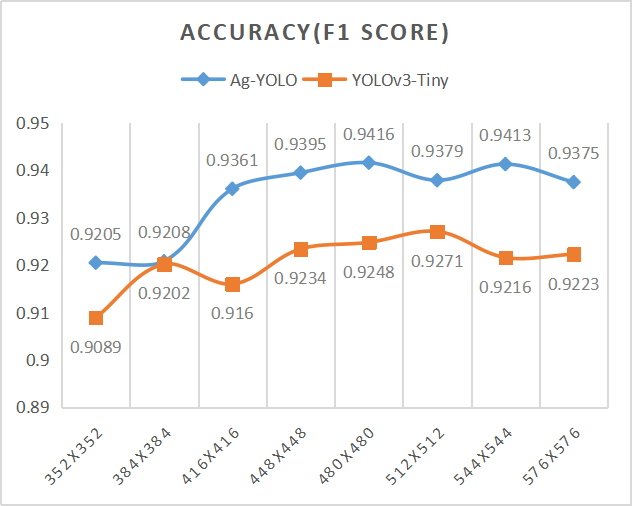}\quad \includegraphics[width=9cm]{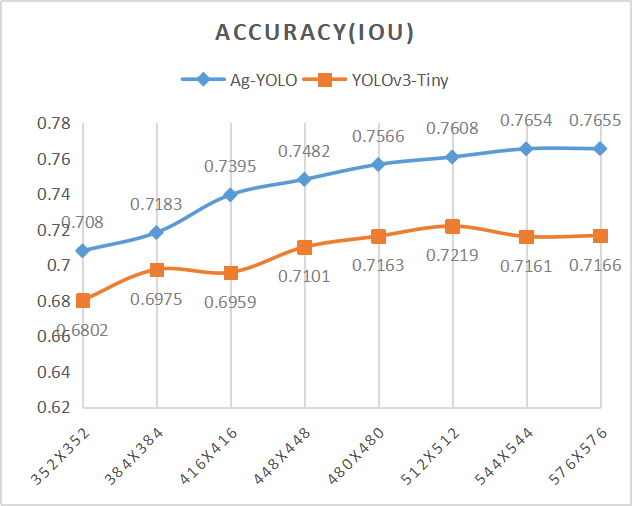}
\caption{Ag-YOLO vs. YOLOv3 tiny under different input dimensions.\label{fig12}}
\end{figure}

\begin{specialtable}[H] 
\caption{Models testd on NCS2. IoU=0.5, input dimensions: 416 $\times$ 416, confidence threshold=0.4 and non-maximum-supress threshold=0.5\label{tab:tab8}} 
\begin{tabular}{p{3cm}ccccp{3.8cm}p{1cm}p{1.1cm}p{1.1cm}}
\toprule
\multirow{2}{*}{\textbf {Model}} & \multicolumn{4}{c}{Features} & \multirow{2}{*}{\textbf {Backbone}} & \multirow{2}{*}{\textbf{FPS} } & \multirow{2}{*}{\textbf{F1 Score}} & \multirow{2}{*}{\textbf{IoU} }\\
\cline{2-5}
& \textbf{CIoU Loss} & \textbf{BG} & \textbf{ResBox} & \textbf{Pruned} & & & &\\
\midrule
1 (YOLOv3-Tiny) & & & & & Darknet18 & 20.7 &0.9160 &0.6959 \\
2 & \checkmark & & & & Darknet18 & 20.7 &0.9276 &0.7148 \\
3 & \checkmark & \checkmark  & & & Darknet18 & 20.7 &0.9302 &0.7253 \\
4 & \checkmark & \checkmark  & \checkmark & & Darknet18 & 26.2 &0.9223 &0.7209 \\
5 & \checkmark & \checkmark  & \checkmark & \checkmark & Darknet18 & 26.3 &0.9209 &0.7108 \\
6 & \checkmark & \checkmark  & \checkmark &\checkmark & PeleeNet & 14.6 &0.9211 &0.7352 \\
7 & \checkmark & \checkmark  & \checkmark & \checkmark & Compact MobileNet v2 & 13.0 &0.9364 &0.7410 \\
8 (Ag-YOLO) & \checkmark & \checkmark  & \checkmark & \checkmark & ShuffleNet v2 derived &\textbf{26.9} &0.9361 &0.7395 \\
\bottomrule
\end{tabular}
\end{specialtable}
\noindent{\textbf{Different input dimensions} Performance of a model is also affected by the input dimension. As in the training phase, we change the input dimensions to $352\times 352$, $384\times 384$, $416 \times 416$, $448\times 448$, $480\times 480$, $512 \times 512$, $544\times 544$, $576\times 576$. }
Figure \ref{fig12} shows performance trends of Ag-YOLO and YOLOv3 tiny.

\begin{figure}[H] 
\centering
\includegraphics[width=9cm]{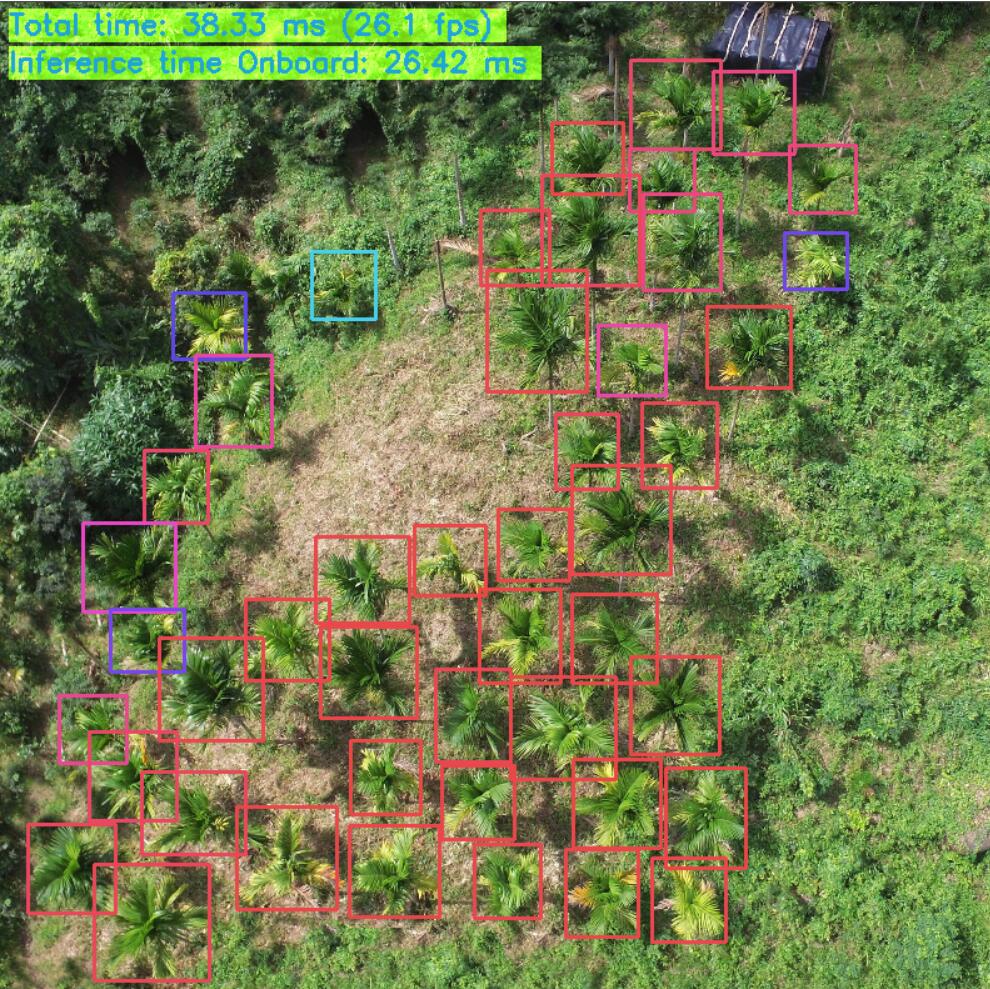}\quad \includegraphics[width=9cm]{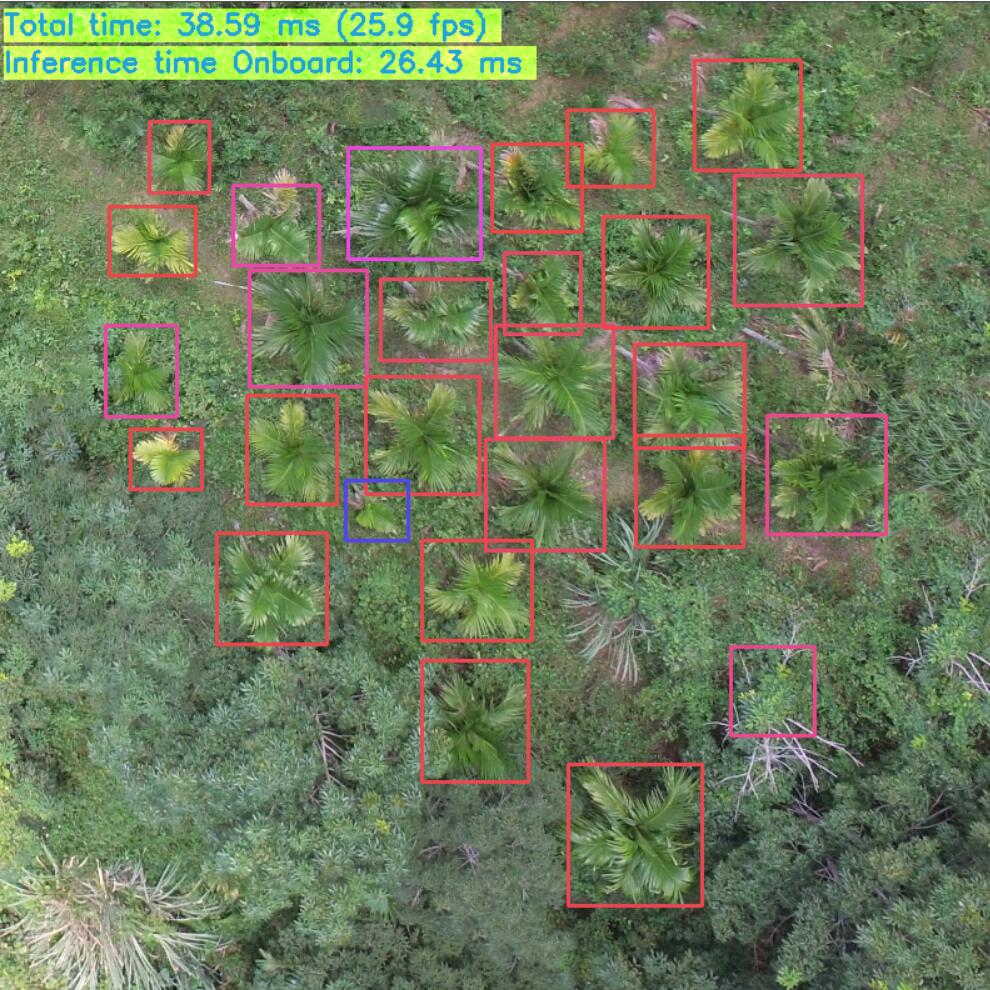}

\noindent{a) R:92\%, P:100\%, F1:0.9583, IoU:0.7199\quad\quad\quad\quad\quad\quad\quad\quad b) R:92.86\%, P:96.30\%, F1:0.9455, IoU:0.7345 }

\caption{Ag-YOLO run on NCS2(Input dimension: 416 $\times$ 416). Images are taken by a UAV, different colors of the predicted square imply different confidence values, blue is low, and red is high\label{fig13}}
\end{figure} 

\section{Discussion}

At Present, more and more researches are done regarding pesticide spraying with UAVs. Lots of the researches in high volume cultivars like tree fruit orchards are dealing with the quality of the spray liquid deposition (\cite{pan2017droplet,tang2018effects,gao2019development,hou2019optimization,martinez2020spray,meng2020experimental}) on the crop, because drift is an important question. A further important research task especially in fruit tree orchards or palm plantations is the identification of the target for only spraying the objects, which have to protect against disease infection. Díaz-Varela, R.A et al. \cite{diaz2015high} used high-resolution UAV Images to access olive tree crown parameters by 3D image processing. Gao, P. et al. \cite{gao2019development} applied a machine learning approach to recognized spray areas from unmanned aerial vehicles. Weiss, M. and Baret, F. \cite{weiss2017using} used 3D point clouds from RGB-images to describe vineyard. Hobart, M. et al.\cite{hobart2020growth} detected growth height of tree walls in apple orchards. If a variable rate spraying via UAV is the aim, (Lian, Q. et al. \cite{lian2019design}, Wen, S. et al.\cite{wen2019design}) the mentioned work in the literature, which are detecting the crown parameters, are important.
Regarding plant diseases like in the present research - the yellow leaf disease of the betel-nut palm – an online algorithm was in the focus of research. Beginning from only a few infected plants diseases can spread over the entire field very fast. Therefore, detecting the target and spraying should be in one operation cycle. This requires embedded camera systems where imaging is followed by an imitate processing. The developed algorithms have to be fast in computational time and simple with few memory capacities. The developed target-detecting algorithm can fulfill this requirement.

The source code of this work is available at \url{https://github.com/rossqin/RQNet}. It can be used as a reference for beginning researchers to develop their real-life AI applications instead of pursuing higher performance with new algorithms with the ever-increasing demand for higher computational power and memory requirements. In a specific agricultural computer vision task, for instance, object detection, the object category is usually one or few, hence it is possible to use a small efficient DNN-based model to achieve a good result. In this work, we prove this by exploring the YOLOv3 tiny architecture, replacing the neck and backbone with different state-of-art efficient DNNs, such as SqueezeNet, MobileNet v2, and ShuffleNet v2. This paper also uses Network Slimming to compress the models to get smaller models. We train all the models on a laptop and test them on a low-cost hardware accelerator, the Intel NCS2. Our architecture, the Ag-YOLO, comprised of a ShuffleNet v2-derived backbone, a ResBlock neck, and a YOLOv3 head, with only 813k parameters and 1.033 billion FLOPs, which is only 9.4\% and 19\% of the darknet18 version respectively, however, brings better accuracy and inference time performance, on the resource-constraint hardware NCS2, we achieve 36.5 fps.
As a camera usually takes video at a frame rate of 24fps, so this is a REAL-TIME object detector.
On the other hand, a compact version of the MobileNet v2 backbone leads to a better accuracy performance, although it takes more than twice BFLOPs and inference time. In a UAV or UGV auto-pilot use case, the host usually moves quite slow, hence we don’t have to process every frame from the on-board camera. For tasks that emphasize accuracy, the compact version of the MobileNet v2 backbone is a better option for Ag-YOLO. To get better accuracy, redundant information between successive frames can be utilized as in\cite{bozek2018towards}. 

It should be beard in mind, that not only RGB cameras attached to UAVs are used for the detection of relatively large target objects (compared to field crops like cereals) or rather disease-infected trees in orchards or plantations. Abdulridha, J. et al. \cite{abdulridha2019uav} applied a hyperspectral camera for detecting citrus canker disease in citrus plantations. Modica, G. et al.\cite{modica2020monitoring} used UAV multi-spectral imagery to monitor the vigor in heterogeneous citrus and olive orchards. Ye, H. et al. \cite{ye2020identification} identify Fusarium wilt in banana using supervised classification algorithms with UAV-based multi-spectral imagery. Those camera systems are expensive, not easy to operate, relatively large, and susceptible in crash situations compared to RGB-cameras.

\section{Conclusions}

By decomposing the fastest object detection algorithm "YOLOv3" into a backbone network which extracts features from a picture, one or more necks which synthesize features the backbone network outputs, and transform them into object categories and locations in different scales, and corresponding heads which decode the information as required, and improving them via more efficient backbones and better neck structure, and taking advantage of some "Freebies" and "Back-of-Specials" such as CIOU Loss and more prior boxes in heads, this paper demonstrated that a DNN-based CV algorithm can be implemented on resource-constraint device to deal with real-life precision agriculture challenge, even with the most cost-efficient embedded AI device, the NCS2, our Ag-YOLO can achieve 36.5fps with satisfying accuracy. 

This experiment also demonstrated that a MobileNetv2-derived backbone has better representational power yet a ShuffleNetv2-like backbone runs faster at the cost of a little accuracy degrade, and both of them are superior in computation intensity and memory usage. With this work including the open-source toolset, it should be very easy to make their legacy agricultural machinery intelligent with an on-board camera and an edge computing device.

\authorcontributions{Conceptualization, Wang, W.; methodology, Qin, Z.Q.; software, Qin, Z.Q.; writing---original draft preparation, Qin, Z.Q.; writing---review and editing, Dammer, K.-H., Qin, Z.W., Guo, L. and Cao, Z.; data curation, Qin, Z.W.; formal analysis, Qin, Z.W.; project administration, Wang, W. and Guo, L.; funding acquisition, Guo, L.; proofread, Cao, Z.. All authors have read and agreed to the published version of the manuscript.}

\funding{National Natural Science Foundation of China(Grant No. 31860180, 32060321); Major Science and Technology Program of Inner Mongolia Autonomous Region (Grant No. ZD20190039)}
 
\conflictsofinterest{The authors declare no conflict of interest.}

\section*{References}

\externalbibliography{yes}

\bibliography{Resources/references.bib} 

\end{document}